\documentclass{new_tlp}
\usepackage{mathptmx}
\usepackage{amssymb}
\usepackage{amsmath}    
\usepackage{graphicx}
\usepackage{url, multicol} 
\usepackage{verbatim}
\usepackage{color} 

\usepackage[normalem]{ulem} 

\newtheorem{definition}{Definition}

\newtheorem{proposition}{Proposition}

\newtheorem{example}{Example}
\newtheorem{lemma}{Lemma}

\begin{document}

\newcommand{\hide}[1]{}
\newcommand{\otherquestions}[1]{}
\newcommand{\catom}[2]{(\{#1\}, \{#2\})}
\newcommand{\set}[1]{\{#1\}}
\newcommand{\pg}[1]{{\tt #1}}
\newcommand{\emptyclause}{\Box}

\def\st{\bigskip\noindent}

\newcommand{\kleene}{\wedge}

\newcommand{\ee}[1] {
  \begin{enumerate}
    #1 
  \end{enumerate}
}

\newcommand{\ie}[1] {
  \begin{enumerate}
    #1 
  \end{enumerate}
}

\newcounter{statement}
\newcommand{\ml}[1]{
    \refstepcounter{statement}
    (\arabic{statement}) 
    \label{#1}
}
\newcommand{\mr}[1]{(\ref{#1})}

\author[Gelfond and Zhang]
         {Michael Gelfond and Yuanlin Zhang\\
          Texas Tech University, Lubbock, Texas 79414, USA\\
         \email{\{michael.gelfond, y.zhang\}@ttu.edu}}

\title{Vicious Circle Principle and Logic Programs with Aggregates}

\maketitle 

  \begin{abstract}
    The paper presents a knowledge representation language
    $\mathcal{A}log$ which extends ASP with aggregates. The goal is to
    have a language based on simple syntax and clear intuitive and
    mathematical  semantics. We give some properties of
    $\mathcal{A}log$, an algorithm for computing its answer sets, and
    comparison with other approaches.

  \end{abstract}


\section{Introduction}
The development of answer set semantics for logic programs \cite{gl88,gl91} led 
to the creation of powerful knowledge representation language,
Answer Set Prolog (ASP), capable of representing recursive definitions,
defaults, effects of actions and other important phenomena of natural 
language. The design of algorithms for computing answer sets and their 
efficient implementations in systems called \emph{ASP solvers} \cite{nss02,dlv06,clasp07} allowed 
the language to become a powerful tool for building 
non-trivial knowledge intensive applications  
\cite{BrewkaET11,ErdemLL12}. 
There are a 
number of extensions of the 
ASP
which  also
contributed to this success. This paper is about one such extension --
\emph{logic programs with aggregates}. By \emph{aggregates we mean functions 
defined on sets of objects of the domain}. (For
simplicity of exposition we limit our attention to aggregates
defined on finite sets.)
Here is a typical example.
\begin{example}[Classes That Need Teaching Assistants]\label{e1}
Suppose that we have a complete list of students enrolled in 
a class $c$ that is represented by the following collection of 
atoms:
\begin{verbatim}
enrolled(c,mike). 
enrolled(c,john). 
... 
\end{verbatim}
Suppose also that we would like to define a new relation $need\_ta(C)$
that holds iff the class $C$ needs a teaching assistant. In this 
particular school $need\_ta(C)$ is true iff the number of students 
enrolled in the class is greater than $20$. The definition can be 
given by a simple rule in the language of logic programs with 
aggregates:
$$need\_ta(C) \leftarrow card\{X : enrolled(C,X)\} > 20$$
where $card$ stands for the cardinality function. Let us call the 
resulting program $P_0$. 
\end{example}
The program is simple, has a clear intuitive meaning, and can be run 
on some of the existing ASP solvers. However, the situation is more 
complex than that. 
Unfortunately,
currently there is no \emph{the} language of logic programs with 
aggregates. Instead there is a comparatively large collection of such 
languages with different syntax and, even more importantly, different 
semantics \cite{pdb07,nss02,SonP07,FaberPL11,gel02,KempS91}.
As an illustration consider the following example:
\begin{example}\label{e2}

Let $P_1$ consist of the following rule:
$$p(a) \leftarrow card\{X : p(X)\} = 1.$$
Even for this seemingly simple program, there are different opinions 
about its meaning. According to \cite{FaberPL11} the program has one answer 
set $A = \{\ \}$; according to \cite{gel02,KempS91} it has two answer sets: $A_1=\{\ \}$
and $A_2=\{p(a)\}$. 
\end{example}
In our judgment this and other similar ``clashes of intuition'' cause 
a serious impediment to the use of aggregates for knowledge 
representation and reasoning. 
In this paper we aim at addressing this problem by suggesting yet another logic 
programming language with aggregates, called $\mathcal{A}log$, which is based on the 
following design principles:
\begin{itemize}
\item the language should have a simple syntax and intuitive semantics 
  based on understandable informal principles, and 
\item  the informal semantics should have clear and elegant 
  mathematics associated with it. 
\end{itemize}
In our opinion existing extensions of ASP by aggregates 
often do not have clear intuitive principles underlying the semantics 
of the new constructs. 
Moreover, some of these languages violate such original foundational 
principles of ASP as the rationality principle. 
The problem is compounded by the fact 
that some of the semantics of aggregates use rather non-trivial 
mathematical constructions which makes it difficult to understand and explain their 
intuitive meaning. 


The semantics of $\mathcal{A}log$ is based on \emph{Vicious Circle 
  Principle} (VCP):  \emph{no object or property can be introduced by the 
  definition referring to the totality of objects satisfying this 
  property}. 
According to Feferman \cite{fef} the principle was first formulated by 
Poincare \cite{poin1906} in his analysis of paradoxes of set theory. 
Similar ideas were already successfully used in a collection of logic 
programming definitions of stratification including that of stratified 
aggregates (see, for instance, \cite{FaberPL11}. Unfortunately, 
limiting the language to stratified aggregates
eliminates some of the useful forms of circles (see Example \ref{e9} below). 
In this paper we give a new form of VCP which goes beyond stratification: 
 \emph{$p(a)$ cannot be introduced by the 
  definition referring to a set of objects satisfying $p$ if this set can contain $a$.} 
Technically, the principle is incorporated in our new 
definition of answer set (which coincides with the original definition for programs without aggregates). 
The definition is short and 
simple. We hope that, combined with a number of informal examples, it 
will be sufficient for developing an intuition necessary for the use 
of the language. 
The paper is organized as follows. 
In Section~\ref{alog}, we define the syntax and semantics of 
$\mathcal{A}log$.  We give some properties of $\mathcal{A}log$ programs in Section~\ref{prop} 
and present an algorithm for computing an 
answer set of an $\mathcal{A}log$ program in Section~\ref{solver}. 
A comparison with the existing work is done in Section~\ref{comp}, 
and we conclude the paper in Section~\ref{conclusion}. 

\section{Syntax and Semantics of $\mathcal{A}log$}\label{alog}
We start with defining the syntax and intuitive semantics of the language. 
\subsection{Syntax}
Let $\Sigma$ be a (possibly sorted) signature with a finite collection 
of predicate, function, and object constants and $\mathcal{A}$ be a finite 
collection of symbols used to denote functions from finite sets of 
terms of $\Sigma$ into integers. Terms and literals over signature 
$\Sigma$ are defined as usual and referred to as \emph{regular}. 
Regular terms are called \emph{ground} if they contain no variables 
and no occurrences of symbols for arithmetic functions. Similarly for 
literals. 
An \emph{aggregate term} is an expression of 
the form 
\begin{equation}\label{agg-term}
f\{\bar{X}:cond\}  
\end{equation}
where $f \in \mathcal{A}$, $cond$ is a collection of regular 
literals, and $\bar{X}$ is a list of variables occurring in $cond$. 
We refer to an expression
\begin{equation}\label{set-name}
\{\bar{X}:cond\}  
\end{equation}
as a \emph{set name}.
An occurrence of a variable from $\bar{X}$ in (\ref{set-name}) 
is called \emph{bound} within (\ref{set-name}).
If the condition from 
(\ref{set-name}) contains no variables except those in $\bar{X}$ then 
it is read as \emph{the set of all objects of the 
  program satisfying $cond$}.  If $cond$ contains other 
variables, say $\bar{Y}=\langle Y_1,\dots,Y_n\rangle$, then 
$\{\bar{X}:cond\}$ defines the function mapping possible values 
$\bar{c} = \langle c_1,\dots,c_n\rangle$ of these variables into sets 
$\{\bar{X}:cond|^{\bar{Y}}_{\bar{c}}\}$ where 
$cond |^{\bar{Y}}_{\bar{c}}$ is the result of replacing 
$Y_1,\dots,Y_n$ by $c_1,\dots,c_n$. 

\noindent
By an \emph{aggregate atom} we mean an 
expression of the form 
\begin{equation}\label{agg-atom}
\langle aggregate\_term \rangle \langle arithmetic\_relation \rangle 
\langle arithmetic\_term  \rangle 
\end{equation}
where $arithmetic\_relation$ is $>, \geq, <, \leq,=$ or !=,  and  
$arithmetic\_term$ is constructed from variables and integers using 
arithmetic operations, $+$, $-$, $\times$, etc. 

\noindent 
By \emph{e-literals} we mean regular literals possibly preceded by 
default negation $not$. The latter (former) are called \emph{negative}
(\emph{positive}) e-literals.

\noindent
A \emph{rule} of 
$\mathcal{A}log$ is an expression of the form 
\begin{equation}\label{rule}
head \leftarrow pos,neg,agg 
\end{equation}
where $head$ is a disjunction of regular literals, $pos$ and $neg$ are 
collections of regular literals and regular literals preceded by 
$not$ respectively, and $agg$ is a collection of aggregate atoms. 
All parts of the rule, including $head$, can be empty. 
An occurrence of a variable in (\ref{rule}) not bound within any set name
in this rule is called 
\emph{free} in (\ref{rule}). 
A rule of $\mathcal{A}log$ is called \emph{ground} 
if it contains no occurrences of free variables and no occurrences of arithmetic functions.

 \st 
A \emph{program} of $\mathcal{A}log$ is a finite collection of 
$\mathcal{A}log$'s rules. A program is \emph{ground} if its rules 
are ground. 

\medskip 
As usual for ASP based languages, 
rules of $\mathcal{A}log$ program with variables are 
viewed as collections of their ground instantiations. 
A \emph{ground instantiation} of a rule $r$ is the program 
obtained from $r$ by 
replacing free occurrences of variables in $r$ by 
ground terms of $\Sigma$ and evaluating all arithmetic functions. 
If the signature $\Sigma$ is sorted (as, for 
instance, in \cite{BalaiGZ13}) the substitutions should respect sort 
requirements for predicates and functions.

\noindent
Clearly the grounding of an
$\mathcal{A}log$ program is a ground program. 
The following examples 
illustrate the definition:

\begin{example}[Grounding: all occurrences of the set variable are bound] \label{e3}
Consider a program $P_2$ with variables:

\begin{verbatim}
q(Y) :- card{X:p(X,Y)} = 1, r(Y). 
r(a).  r(b).  p(a,b). 
\end{verbatim}
Here all occurrences of a set variable $X$ are bound; all occurrences
of a variable $Y$ are free.
The program's grounding, $ground(P_2)$, is 

\begin{verbatim}
q(a) :- card{X:p(X,a)} = 1, r(a). 
q(b) :- card{X:p(X,b)} = 1, r(b). 
r(a).  r(b).  p(a,b). 
\end{verbatim}
\end{example}
The next example deal with the case when some occurrences of the set
variable in a rule are free and some are bound.  

\begin{example}[Grounding: some occurrences of a set variable are free]\label{e4}
Consider an $\mathcal{A}log$ program $P_3$
\begin{verbatim}
r :- card{X:p(X)} >= 2, q(X). 
p(a).  p(b).  q(a). 
\end{verbatim}
Here the occurrence of $X$ in $q(X)$ is free. Hence the ground program 
$ground(P_3)$ is:
 \begin{verbatim}
r :- card{X:p(X)} >= 2, q(a). 
r :- card{X:p(X)} >= 2, q(b). 
p(a).  p(b).  q(a). 
\end{verbatim}
\end{example}

\noindent
Note that despite its apparent simplicity the syntax of 
$\mathcal{A}log$ differs substantially from syntax of most other logic 
programming languages allowing aggregates 
(with the exception of that in \cite{gel02}).
We illustrate the differences using the language presented in 
\cite{FaberPL11}. (In what follows we refer to this language as $\mathcal{F}log$.) 
While syntactically programs of $\mathcal{A}log$ can also be viewed as 
programs of $\mathcal{F}log$ the opposite is not true. Among other 
things $\mathcal{F}log$ allows parameters of aggregates to be 
substantially more complex than those of $\mathcal{A}log$. For 
instance, an expression $f\{a:p(a,a),b:p(b,a)\} = 1$ where $f$ is an 
aggregate atom of $\mathcal{F}log$ but not of 
$\mathcal{A}log$. This construction which is different from a usual 
set-theoretic notation used in $\mathcal{A}log$ is important for the 
$\mathcal{F}log$ definition of grounding. 
For instance the 
grounding of the first rule of program $P_2$ from Example \ref{e3} 
understood as a program of $\mathcal{F}log$ consists of $\mathcal{F}log$
rules 
\begin{verbatim}
q(a) :- card{a:p(a,a),b:p(b,a)} = 1, r(a). 
q(b) :- card{a:p(a,b),b:p(b,b)} = 1, r(b). 
\end{verbatim}
which is not even a program of $\mathcal{A}log$. 
Another important difference between the grounding methods of these 
languages can be illustrated by the $\mathcal{F}log$ grounding $ground_f(P_3)$ of 
program $P_3$ from Example \ref{e4} that looks as follows:
\begin{verbatim}
r :- card{a:p(a)} >= 2, q(a). 
r :- card{b:p(b)} >= 2, q(b). 
p(a).  p(b).  q(a). 
\end{verbatim}
Clearly this is substantially different from the $\mathcal{A}log$ grounding of 
$P_3$ from Example \ref{e4}. In Section \ref{comp} we show that this 
difference in grounding reflects substantial semantic differences 
between the two languages.

\subsection{Semantics}
To define the semantics of $\mathcal{A}log$ programs we expand the 
standard definition of answer set from \cite{gl88}. The resulting 
definition captures the rationality principle - \emph{believe 
  nothing you are not forced to believe} \cite{GelK13} - and avoids vicious circles. 
As usual the definition of answer set is given for ground 
programs. 
Some terminology:
a ground aggregate atom $f\{X:p(X)\} \odot  n$
(where $\odot$ is one of the arithmetic relations allowed in the language) 
is \emph{true} in a set of ground regular literals $S$ if $f\{X: p(X) 
\in S\} \odot  n$; otherwise the atom is \emph{false} in $I$. 
\begin{definition}[Aggregate Reduct]\label{reduct}
The \emph{aggregate reduct} of a ground program $\Pi$ 
of $\mathcal{A}log$ with respect to a set of 
ground regular literals $S$ is obtained from $\Pi$ by 
\begin{enumerate}
\item removing from $\Pi$ all rules containing aggregate atoms false 
in $S$. 
\item replacing every remaining aggregate atom $f\{X:p(X)\} \odot  n$ by 
  the set $\{p(t) : p(t) \in S\}$ (which is called the \emph{reduct of 
    the aggregate} with respect to $S$). 
\end{enumerate}
\end{definition}
(Here $p(t)$ is the result of replacing variable $X$ by ground term $t$). The 
second clause of the definition reflects the principle of avoiding 
vicious circles -- a rule with aggregate atom $f\{X:p(X)\} \odot  n$
in the body can only be used if ``the totality'' of all objects satisfying 
$p$ has already being constructed. Attempting to apply this rule to 
define $p(t)$ will either lead to contradiction or to turning the rule 
into tautology (see Examples \ref{e7} and \ref{e9}).


\begin{definition}[Answer Set]\label{ans-set}
A set $S$ of ground regular literals over the signature of a ground
program $\Pi$ of $\mathcal{A}log$ is an 
\emph{answer set} of $\Pi$ if it is an answer set of an aggregate 
reduct of $\Pi$ with respect to $S$. 
\end{definition}
We will illustrate this definition by a number of examples. 
\begin{example}[Example \ref{e3} Revisited] \label{e5}
Consider a program $P_2$ and its grounding from Example \ref{e3}. 
It is easy to see that the aggregate reduct of the program with 
respect to any set $S$ not containing $p(a,b)$ consists of the program 
facts, and hence $S$ is not an answer set of $P_2$. However the program's 
aggregate reduct with respect to 
$A = \{q(b),r(a),r(b),p(a,b)\}$ 
consists of the program's facts and the rule 
\begin{verbatim}
q(b) :- p(a,b),r(b).  
\end{verbatim}
Hence $A$ is an answer set of $P_2$. 
\end{example}

\begin{example}[Example \ref{e4} Revisited] \label{e6}
Consider now the grounding 
 \begin{verbatim}
r :- card{X:p(X)} >= 2, q(a). 
r :- card{X:p(X)} >= 2, q(b). 
p(a).  p(b).  q(a). 
\end{verbatim}
of program $P_3$ from Example \ref{e4}. 
Any answer set $S$ of this program must contain its facts. Hence 
$\{X: p(X) \in S\} = \{a,b\}$. $S$ satisfies the body of the first 
rule and must also contain $r$. Indeed, the aggregate reduct of $P_3$
with respect to $S = \{p(a), p(b), q(a), r\}$ consists of the facts of 
$P_3$ and the rules 
\begin{verbatim}
r :- p(a),p(b),q(a). 
r :- p(a),p(b),q(b). 
\end{verbatim}
Hence $S$ is the answer set of $P_3$. 
\end{example}
Neither of the two examples above required the application of VCP.
The next example shows how this 
principle influences our definition of answer sets and hence our reasoning. 
\begin{example}[Example \ref{e2} Revisited]\label{e7}
Consider a program $P_1$ from Example \ref{e2}. 
The program, consisting of a rule 
\begin{verbatim}
p(a) :- card{X : p(X)}=1 
\end{verbatim}
is grounded. It has two candidate answer sets, $S_1 = \{\ \}$ and $S_2 
= \{p(a)\}$. The aggregate reduct of the program with respect to $S_1$
is the empty program. Hence, $S_1$ is an answer set of $P_1$. 
The program's aggregate reduct with respect to $S_2$ however is 
\begin{verbatim}
p(a) :- p(a). 
\end{verbatim}
The answer set of this reduct is empty and hence $S_1$ is the only 
answer of $P_1$. 
\end{example}
Example \ref{e7} shows how the attempt to define $p(a)$ in terms of 
totality of $p$ turns the defining rule into a tautology. 
The next example shows how it can lead to inconsistency of a program. 
\begin{example}[Vicious Circles through Aggregates and 
  Inconsistency] \label{e8}
Consider a program $P_4$:
\begin{verbatim}
p(a). 
p(b) :- card{X:p(X)} > 0. 
\end{verbatim}
Since every answer set of the program must contain $p(a)$,
the program has two candidate answer sets: $S_1 = \{p(a)\}$ and 
$S_2 = \{p(a),p(b)\}$. The aggregate reduct of $P_4$ with respect to 
$S_1$ is 
\begin{verbatim}
p(a). 
p(b) :- p(a). 
\end{verbatim}
The answer set of the reduct is $\{p(a),p(b)\}$ and hence $S_1$ is not 
an answer set of $P_4$. The reduct of $P_4$ with respect to $S_2$ is 
\begin{verbatim}
p(a). 
p(b) :- p(a),p(b). 
\end{verbatim}
Again its answer set is not equal to $S_2$ and hence $P_4$ is 
inconsistent (i.e., has no answer sets). The inconsistency is the direct 
result of an attempt to violate the underlying principle of the 
semantics. 
Indeed, the definition of $p(b)$ refers to the set of objects
satisfying $p$ that can contain $b$ which is prohibited by our version
of VCP.
One can, of course, argue that $S_2$ can be viewed as 
a reasonable collection of beliefs which can be formed by a rational 
reasoner associated with $P_4$. After all, we do not need the totality 
of $p$ to satisfy the body of the rule defining $p(b)$. It is 
sufficient to know that $p$ contains $a$. This is indeed true but 
this reasoning depends on the knowledge which is not directly 
incorporated in the definition of $p(b)$. 
If one were to replace $P_4$ by 
\begin{verbatim}
p(a). 
p(b) :- card{X:p(X), X != b} > 0. 
\end{verbatim}
then, as expected, the vicious circle principle will not be violated 
and the program will have unique answer set $\{p(a),p(b)\}$. 
\end{example}
We end this section by a simple but practical example of a program 
which allows recursion through aggregates but avoids vicious circles. 
\begin{example}[Defining Digital Circuits]\label{e9}
Consider part of a logic program formalizing propagation of binary signals 
through simple digital circuits. We assume that the circuit does not 
have a feedback, i.e., a wire receiving a signal from a gate cannot be 
an input wire to this gate. 
The program may contain a simple 
rule 
\begin{verbatim}
  val(W,0) :-
        gate(G, and),
        output(W, G),
        card{W: val(W,0), input(W, G)} > 0. 
\end{verbatim}
(partially) describing propagation of symbols through an \emph{and}
gate. Here $val(W,S)$ holds iff the digital signal on a wire $W$ has 
value $S$.  Despite its recursive nature the definition of $val$
avoids vicious circle. To define the signal on an output wire $W$ of an 
\emph{and} gate $G$ one needs to only construct a particular subset of 
input wires of $G$. Since, due to absence of feedback in our circuit,
$W$ can not belong to the latter set our definition is reasonable. 
To illustrate that our definition of answer set produces the intended 
result let us consider program $P_5$ consisting of the above rule and a 
collection of facts:
\begin{verbatim}
gate(g, and). 
output(w0, g). 
input(w1, g). 
input(w2, g).  
val(w1,0). 
\end{verbatim}

\noindent The grounding, $ground(P_5)$, of $P_5$ consists of
the above facts and the three rules of the form
\begin{verbatim}
  val(w,0) :-
        gate(g, and),
        output(w, g),
        card{W: val(W,0), input(W, g)} > 0. 
\end{verbatim}
where $w$ is $w_0$, $w_1$ ,and $w_2$.

\noindent Let $S=\{gate(g, and), val(w1, 0), val(w0,0), output(w0, g), input(w1, g),
  input(w2, g)\}$. 
The aggregate reduct of $ground(P_5)$
with respect to $S$ is the 
collection of facts  and the rules 
\begin{verbatim}
  val(w,0) :-
        gate(g, and),
        output(w, g),
        input(w1, g), 
        val(w1, 0).         
\end{verbatim}
where $w$ is $w_0$, $w_1$, and $w_2$.

\noindent The answer set of the reduct is $S$ and hence $S$ is an 
answer set of $P_5$. As expected it is the only answer set. 
(Indeed it is easy to see that other candidates 
do not satisfy our definition.) 
 
\end{example}

\section{Properties of $\mathcal{A}log$ programs} \label{prop}
In this section we give some basic properties of $\mathcal{A}log$ programs. 
Propositions \ref{p1} and \ref{p2} ensure that, as in regular ASP,
answer sets of $\mathcal{A}log$ program are formed using the program 
rules together with the rationality principle. 
Proposition \ref{split} is the $\mathcal{A}log$ version of the basic 
technical tool used in theoretical investigations of ASP and its extensions. 
Proposition \ref{compl} shows that complexity of entailment in 
$\mathcal{A}log$ is the same as that in regular ASP. 

We will use the following terminology:
e-literals $p$ and $not\ p$ are called \emph{contrary};
$not\ l$ denotes a literal contrary to e-literal $l$;
a \emph{partial interpretation} $I$ over signature $\Sigma$ is a 
consistent set of e-literals of this signature; an e-literal $l$ is \emph{true} 
in $I$ if $l \in I$; it is  \emph{false} if $not\ l \in I$; otherwise 
$l$ is \emph{undefined} in $I$. 
An aggregate atom  $f\{X: q(X)\}\ \odot\  n$ is \emph{true} in $I$ if 
$f\{t: q(t) \in I\}\ \odot\ n$ is true, i.e., the value of $f$ on 
the set $\{t: q(t) \in I\}$ and the number $n$ satisfy property $\odot$. 
Otherwise, the atom is \emph{false} in $I$. 
The head of a rule is \emph{satisfied} by $I$ if at least one of its literals is 
true in $I$; the body of a rule is  \emph{satisfied} by $I$ if all of 
its aggregate atoms and e-literals are true in $I$. 
A \emph{rule is satisfied} by $I$ if its head is satisfied by $I$ or 
its body is not satisfied by $I$. 
\begin{proposition}[Rule Satisfaction and Supportedness]\label{p1} Let 
  $A$ be an answer set of a ground $\mathcal{A}log$ program $\Pi$. 
Then 
\ee{
  \item $A$ satisfies every rule $r$ of $\Pi$. 
  \item If $p \in A$ then there is a rule $r$ from $\Pi$  such 
    that the body of $r$ is satisfied by $A$ and $p$ is the only atom 
    in the head of $r$ which is true in $A$. (It is often said that 
    rule $r$ supports atom $p$.) 
}
\end{proposition}
\begin{proposition}[Anti-chain Property]\label{p2}
Let $A_1$ be an answer set of an $\mathcal{A}log$ program $\Pi$. 
Then there is no answer set $A_2$ of $\Pi$ such that $A_1$ is a proper 
subset of $A_2$. 
\end{proposition}
\begin{proposition}[Splitting Set Theorem]\label{split}
Let $\Pi_1$ and $\Pi_2$ be programs of $\mathcal{A}log$ such that no atom 
occurring in $\Pi_1$ is a head atom of $\Pi_2$. Let $S$ be a set of 
atoms containing all head atoms of $\Pi_1$ but no head atoms of 
$\Pi_2$. A set $A$ of atoms is an answer set of $\Pi_1 \cup \Pi_2$
iff $A \cap S$  is an answer set of $\Pi_1$ and $A$ is an answer set 
of $(A \cap S) \cup \Pi_2$. 
\end{proposition}
\begin{proposition}[Complexity]\label{compl}
The problem of checking if a ground atom $a$ belongs to all answer 
sets of an $\mathcal{A}log$ program is $\Pi_2^P$ complete. 
\end{proposition}
\section{An Algorithm for Computing Answer Sets}\label{solver}
In this section we briefly outline an algorithm, called $\mathcal{A}solver$,
for computing answer 
sets of $\mathcal{A}log$ programs. 
We follow the tradition and limit our attention to programs 
without classical negation. Hence, in this section we consider
only programs of this type.
By an \emph{atom} we mean an e-atom or an aggregate atom.

\begin{definition}[Strong Satisfiability 
  and Refutability]\label{strong-satisfiablity} 
\begin{itemize}
\item An atom is \emph{strongly satisfied} (\emph{strongly refuted}) 
by a partial 
interpretation $I$ if it is true (false) in every partial interpretation 
containing $I$; an atom which is neither strongly satisfied nor 
strongly refuted 
by $I$ is \emph{undecided} by $I$. 

\item A set $S$ of atoms is \emph{strongly satisfied} 
by $I$ if all atoms in $S$ are strongly satisfied by $I$; 

\item $S$ is  \emph{strongly 
  refuted} by $I$ if for every 
  partial interpretation $I^\prime$ containing $I$, some atom of $S$ is false 
in $I^\prime$. 
\end{itemize}

\end{definition}

\noindent 
For instance, an e-atom is strongly satisfied (refuted) by $I$ iff it is true 
(false) in $I$;
an atom  $card\{X:p(X)\} > n$ which is true in $I$ is 
strongly satisfied by $I$; an atom $card\{X:p(X)\} < n$ which is false in $I$ is 
strongly refuted by $I$; and a set $\{f\{X: p(X)\} > 5, f\{X:
p(X)\} < 3\}$ is strongly refuted by any partial interpretation. 

$\mathcal{A}solver$ consists of three functions: $Solver$, $Cons$, and 
$IsAnswerSet$. 
The main function, $Solver$, is similar to that used in standard ASP 
algorithms (See, for instance, $Solver1$
from \cite{GelK13}). But unlike these functions which normally have two 
parameters - partial interpretation $I$ and program $\Pi$ - $Solver$
has two additional parameters, $TA$ and $FA$ containing aggregate atoms 
that must be true and false respectively in the answer set 
under construction. $Solver$ returns $\langle I, true\rangle$
where $I$ is an answer
set of $\Pi$ compatible with its parameters and $false$ if no such
answer set exists.
The $Solver$'s description will be omitted due to 
space limitations. 
The second function, $Cons$, computes the consequences of 
its parameters - a program $\Pi$, a partial interpretation $I$, and two 
above described sets $TA$ and $FA$ of aggregates atoms. 
Due to the presence of aggregates the function is sufficiently 
different from a typical $Cons$ function of ASP solvers so we 
describe it in some detail. The new value of $I$, containing 
the desired consequences is computed 
by application of the following {\bf inference rules}: 

\begin{enumerate}
  \item If the body of a rule $r$ is strongly satisfied by $I$ and all 
    atoms in the head of $r$ except $p$ are false in $I$ then $p$ must be in $I$.   
  \item If an atom $p \in I$ belongs to the head of exactly one rule $r$
    of $\Pi$ then every other atom from the head of $r$ must have its 
    complement in $I$, the e-atoms from the body of $r$ must be in 
    $I$ and its aggregate atoms must be in $TA$. 
  \item If every atom of the head of a rule $r$ is false in $I$, and $l$ is 
the only premise of $r$ which is either an undefined e-atom or 
an aggregate atom not in $FA$, and  the rest of the body is strongly 
satisfied by $I$, then 
  \ee{
    \item if $l$ is an e-atom, then the complement of $l$ must be in $I$,
	\item if $l$ is an aggregate atom, then it must be in $FA$. 
  }
  \item \label{firstSupportRule}   
  If the body of every rule with $p$ in the head is strongly refuted by $I$, then $(not\ p)$ must be in $I$. 
\end{enumerate}


\st Given an interpretation $I$, a program $\Pi$, inference
rule $i \in [1..4]$ and $r \in \Pi$, let function $iCons(i,I,\Pi,r)$ 
return $<\delta I, \delta TA, \delta FA>$ where 
$\delta I$, $\delta TA$ and $\delta FA$ are the results of applying
inference rule $i$ to $r$. (Note, that inference rule $4$ does not 
really use $r$). We also need the following terminology.
We say that $I$ is \emph{compatible} with $TA$ if  $TA$
is not strongly refuted by $I$; $I$ is \emph{compatible} with $FA$
if no atom from $FA$ is strongly satisfied by $I$. 
A set $A$ of regular atoms is 
\emph{compatible} with $TA$ and $FA$ if the set 
$compl(A) = \{p : p \in A\} \cup 
\{not~ a: a \notin A\}$ is \emph{compatible}
with $TA$ and $FA$; $A$ is compatible with $I$ if $I \subseteq 
compl(A)$. 
The algorithm $Cons$ is listed below.


\begin{tabbing}
abcde\=ab\=ab\=ab\aa\=aa\=\kill 
{\bf function} Cons \\
\> {\bf input}: \>\>\>partial interpretation $I_0$, sets $TA_0$ and $FA_0$
of aggregate atoms compatible with $I_0$, \\
\> \> \> \> and program $\Pi_0$ with signature $\Sigma_0$; \\
\> {\bf output}: \\
\> \> $\langle \Pi, I, TA, FA, true\rangle$ where $I$ is a partial interpretation such that $I_0 \subseteq I$, \\ 
\> \> \> \> $TA$ and $FA$ are sets of aggregate atoms such that $TA_0 \subseteq TA$ and $FA_0 \subseteq FA$, \\
\> \> \> \> $I$ is compatible with $TA$ and $FA$, and $\Pi$ is a program with signature $\Sigma_0$ such that \\
\> \> \> \> for every A, \\
\> \> \> \> \> $A$ is an answer set of $\Pi_0$ that is compatible with $I_0$ iff  $A$ is an answer set of $\Pi$ \\
\> \> \> \> \> that is compatible with $I$. \\
\> \> $\langle\Pi_0, I_0, TA_0, FA_0, false\rangle$ if there is no answer set of $\Pi_0$ compatible with $I_0$; \\
\> {\bf var} $I, T$: set of e-atoms; $TA, FA$: set of aggregate atoms; $\Pi$: program; \\
1. \> Initialize $I$, $\Pi$, $TA$ and $FA$ to be $I_0$, $\Pi_0$, $TA_0$ and $FA_0$ respectively; \\
2. \> {\bf repeat} \\
3. \> \> $T$ := $I$; \\
4. \> \> Remove from $\Pi$ all the rules whose bodies are strongly falsified by $I$; \\
5. \> \> Remove from the bodies of rules of $\Pi$ \\
   \> \> \> all negative e-atoms true in $I$ and aggregate atoms strongly satisfied by $I$; \\
6. \> \> Non-deterministically select an inference rule $i$ from (1)--(4); \\
8. \> \> {\bf for} every $r \in \Pi$ \\ 
9. \> \> \> $<\delta I, \delta TA, \delta FA>$ := $iCons(I,\Pi,i,r)$; \\
10. \> \> \> $I$ := $I \cup \delta I$, $TA$ := $TA \cup \delta TA$, $FA$ := $FA \cup \delta FA$; \\
11. \> {\bf until} $I$ = $T$; \\
12. \> {\bf if} $I$ is consistent, $TA$ and $FA$ are compatible with 
$I$  {\bf then} \\
13. \> \> {\bf return} $<\Pi, I, TA, FA, true>$; \\
14. \> {\bf else}  {\bf return} $<\Pi_0, I_0, TA_0, FA_0, false>$; \\
\end{tabbing}

\noindent The third function, $IsAnswerSet$ of our solver 
$\mathcal{A}solver$ checks if interpretation $I$ is an answer set 
of a program $\Pi$. It computes the aggregate reduct of $\Pi$ with 
respect to $I$ and applies usual checking algorithm (see, for 
instance, \cite{KochLP03}).

\begin{proposition}[Correctness of the Solver]
If, given a program $\Pi_0$, a partial interpretation $I_0$, and sets 
$TA_0$ and $FA_0$ of aggregate atoms Solver($I_0, TA_0, FA_0,
\Pi_0$) returns $\langle I, true\rangle$ then $I$ is 
an answer set of $\Pi_0$ compatible 
with $I_0$, $TA_0$ and $FA_0$. If there is no such answer set,
the solver returns $false$.
\end{proposition}
To illustrate the algorithm consider a program $\Pi$
\begin{verbatim}
:- p(a).
p(a) :- card{X:q(X)} > 0.
q(a) or p(b).
\end{verbatim}
and trace $Solver(\Pi,I,TA,FA)$ where $I$, $TA$, and $FA$ are empty.
$Solver$ starts by calling $Cons$ which computes the consequence $not\
p(a)$ (from the first rule of the program), $FA=\{card\{X:q(X)\} > 0\}$
(from the second rule of the program) and $not\ q(b)$ (from the fourth
inference rule), and returns $true$, $I=\{not\ q(b), not\ p(a)\}$ and
new $FA$; $TA$ is unchanged. 
$Solver$ then guesses $q(a)$ to be true, i.e., 
$I=\{not\ q(b), not\ p(a), q(a)\}$, and calls $Cons$ again.  $Cons$ 
does not produce any new consequences but finds that $FA$ is 
not compatible with $I$ (line 12 of the algorithm). So, it returns
$false$, which causes $Solver$ to set $q(a)$ to be false, i.e., 
$I=\{not\ q(b), not\ p(a)$, $not\ q(a)\}$. $Solver$ then calls $Cons$ 
again which returns $I=\{not\ q(b), not\ p(a), not\ q(a), p(b)\}$. 
$Solver$ finds that $I$ is complete and calls $IsAnswerSet$ which returns true. 
Finally, $Solver$ returns $I$ as an answer set of the program. 
\section{Comparison with Other Approaches}\label{comp}
There are a large number of approaches to the syntax and semantics of
extensions of ASP by aggregates. In this section we concentrate on
languages from \cite{SonP07}  and \cite{FaberPL11} which we refer to as
$\mathcal{S}log$
and $\mathcal{F}log$ respectively. Due to multiple equivalence
results discussed in these papers 
this 
is sufficient to cover most of the approaches.
The main difference between the syntax of aggregates in
$\mathcal{A}log$ and $\mathcal{F}log$ is in treatment of variables
occurring in \emph{aggregate terms}. $\mathcal{A}log$ uses usual logical concept of
bound and free occurrence of a variable (the occurrence of $X$ within
$S=\{X:p(X,Y)\}$
is bound while the occurrence of $Y$ is free).
$\mathcal{F}log$ uses very different concepts of global and local
variable of a \emph{rule}. A \emph{variable is local in rule $r$ if it
occurs solely in an aggregate term of $r$; otherwise, the variable is
global}. 
As the result, in $\mathcal{A}log$, every aggregate
term $\{X : p(X)\}$ can be replaced by a term  $\{Y : p(Y)\}$
while it is not the case in $\mathcal{F}log$.
In our opinion the approach of $\mathcal{F}log$ (and many
other languages and systems which adopted this syntax) 
makes
declarative reading of aggregate terms substantially more
difficult\footnote{The other 
difference in reading of $S$ is related to the treatment of
variable $Y$. In $\mathcal{F}log$ the variable is bound by an unseen
existential quantifier. If all the variables are local then
$S=\{X:p(X,Y)\}$ is really $S_1 = \{X: \exists Y \ p(X,Y)\}$. In
$\mathcal{A}log$ $Y$ is free. Both approaches are reasonable but we
prefer to deal with the different possible readings by introducing an explicit
existential quantifier as in Prolog. It is easy semantically and we
do not discuss it in the paper.}.
To see the semantic ramifications of the $\mathcal{F}log$ treatment of variables
consider the following example:

\begin{example}[Variables in Aggregate Terms: Global versus Bound]\label{e10}
Consider program $P_3$ from 
Example \ref{e4}. According to $\mathcal{F}log$ the meaning of an occurrence
of an expression $\{X:p(X)\}$ in the body of the program's first rule changes if $X$ is replaced by
a different variable. In $\mathcal{A}log$, where $X$ is
understood as bound 
this is not the case.
This leads to substantial difference in grounding and in
the semantics of the program. 
In $\mathcal{A}log$ $P_3$ has one answer set, $\{p(a),p(b),q(a),r\}$. 
In $\mathcal{F}log$ answer sets of $P_3$ are those of $ground_f(P_3)$.
The answer set of the latter is $\{p(a),p(b),q(a)\}$. 
\end{example}

Other semantic differences are due to the multiplicity of informal
(and not necessarily clearly spelled out) principles underlying various semantics. 

\begin{example}[Vicious Circles in $\mathcal{F}log$] \label{e11a} 
Consider the following program, $P_6$, adopted from  \cite{SonP07}:
\begin{verbatim}
p(1) :- p(0).
p(0) :- p(1).
p(1) :- count{X: p(X)} != 1. 
\end{verbatim}
which, if viewed as $\mathcal{F}log$ program, has one answer set $A=\{p(0),p(1)\}$. 
Informal argument justifying this result goes something like this:
Clearly, $A$ satisfies the rules of the program. To satisfy the
minimality principle no proper subset of $A$ should be able to do
that, which is easily checked to be true. 
Faber et al use so called \emph{black box 
principle}: ``when checking stability they [aggregate literals] are
either present in their entirety or missing altogether'', i.e.,
the semantics of $\mathcal{F}log$ does not consider
the process of derivation of elements of
the aggregate parameter. Note however, that the program's definition of $p(1)$
is given in terms of fully defined term $\{X:p(X)\}$, i.e., the
definition contains a vicious circle. This explains why $A$ is not an
answer set of $P_6$ in $\mathcal{A}log$.
In this particular example we are in agreement with $\mathcal{S}log$
which requires that the value of an aggregate atom can be computed
before the rule with this atom in the body can be used in the
construction of an answer set.

\end{example}
The absence of answer set of $P_6$ in  $\mathcal{S}log$ may suggest that
it adheres to our formalization of the VCP. 
The next example shows that it is not the case.

\begin{example}[VCP and Constructive Semantics of aggregates]\label{e12}
Let us consider a program $P_7$.
\begin{verbatim}
p(a) :- count{X:p(X)} > 0.
p(b) :- not q.
q :- not p(b).
\end{verbatim}
As shown in \cite{SonP07} the program has two $\mathcal{S}log$ answer sets, $A=\{q\}$ and
$B = \{p(a),p(b)\}$. If viewed as a program of  $\mathcal{A}log$,
$P_7$ will have one answer set, $A$. This happens because the
$\mathcal{S}log$ construction
of $B$  uses knowledge about properties
of the  \emph{aggregate atom} of the first rule; the semantics of
$\mathcal{A}log$ only takes into account the \emph{meaning of the  parameter
of the aggregate
term}. Both approaches can, probably, be successfully defended but, in
our opinion, the constructive semantics has a disadvantage of being less general 
(it is only applicable to non-disjunctive programs), and more complex mathematically.
 \end{example}
A key difference between our algorithm and those in the existing work 
\cite{FaberPLDI08,GebserKKS09} is that the other work needs rather 
involved methods to ground the aggregates while our algorithm does not 
need to ground the aggregate atoms. As a result, the ground program
used by our algorithm may be smaller, and our algorithm is simpler.  

There is also a close connection between the above semantics of
aggregates all of which are based on some notion of a reduct or a 
fixpoint computation and approaches in
which aggregates are represented as special cases of more general
constructs, such as propositional formulas \cite{Ferraris05,HarLY13} and abstract constraint
atoms \cite{Marek04,LiuPST10,WangLZY12} (Our semantics can be easily
extended to the latter).
Some of the existing equivalence results allow us to establish 
the relationship between these approaches and $\mathcal{A}log$.
Others require further investigation.


\section{Conclusion and Future Work}\label{conclusion}

We presented an extension, $\mathcal{A}log$, of ASP which allows for the
representation of and reasoning with aggregates. We believe that the language satisfies
design criteria of simplicity of syntax and formal and informal
semantics. There are many ways in which this work can be continued.
The first, and simplest, step is to expand $\mathcal{A}log$ by
allowing choice rules similar to those of \cite{nss02}. This can be
done in a natural way by combining ideas from this paper and that from
\cite{gel02}. 
We also plan to investigate mapping of
$\mathcal{A}log$ into logic programs with arbitrary propositional formulas.
There are many interesting and, we believe, important questions 
related to optimization of the $\mathcal{A}log$ solver from Section \ref{solver}.
After clarity is reached in this area one will, of course, try to
address the questions of implementation.

\section{Acknowledgment} We would like to thank 
Amelia Harrison, Patrick Kahl, Vladimir Lifschitz, and
Tran Cao Son for useful comments. The authors' work was partially supported by NSF grant
IIS-1018031.

\bibliographystyle{acmtrans} 
\bibliography{biblio}

\section{Appendix}

\makeatletter
\@addtoreset{proposition}{}
\makeatother
\setcounter{proposition}{0}

In this appendix, given an ${\mathcal{A}}log$ program $\Pi$, a set $A$ of literals and a rule $r \in \Pi$, 
we use $\alpha(r, A)$ to denote the rule obtained from $r$ in the aggregate reduct of $\Pi$ with respect to $A$. 
$\alpha(r,A)$ is $nil$, called an {\em empty rule}, if $r$ is discarded in the aggregate reduct. 
We use $\alpha(\Pi, A)$ to denote the aggregate reduct of $\Pi$, i.e.,  $\{\alpha(r, A): r \in \Pi \mbox{ and } \alpha(r, A) \neq nil\}$. 
 
\begin{proposition}[Rule Satisfaction and Supportedness] Let 
  $A$ be an answer set of a ground $\mathcal{A}log$ program $\Pi$. 
Then 
\ee{
  \item $A$ satisfies every rule $r$ of $\Pi$. 
  \item If $p \in A$ then there is a rule $r$ from $\Pi$  such 
    that the body of $r$ is satisfied by $A$ and $p$ is the only atom 
    in the head of $r$ which is true in $A$. (It is often said that 
    rule $r$ supports atom $p$.) 
}
\end{proposition}

\noindent
Proof: Let 

\st (1) $A$ be an answer set of $\Pi$.

\st We first prove $A$ satisfies every rule $r$ of $\Pi$. Let $r$ be a rule of $\Pi$ such that 

\st (2) $A$ satisfies the body of
  $r$. 
  
\st Statement (2) implies that every aggregate atom, if there is any, of the body of $r$ is satisfied by $A$.  By the definition of the aggregate reduct, there must be a non-empty rule
  $r^\prime \in \alpha(\Pi, A)$ such that

\st (3)  $r^\prime = \alpha(r,A)$.  

\st By the definition of aggregate reduct, $A$ satisfies the body of $r$ iff it satisfies that of $r^\prime$. Therefore, (2) and (3) imply that 

\st (4) $A$ satisfies the body of $r^\prime$. 

\st By the definition of answer set of $\mathcal{A}log$, (1) implies that 

\st (5) $A$ is an answer set of $\alpha(\Pi,A)$.

\st Since $\alpha(\Pi,A)$ is an ASP program, (3) and (5) imply that 

\st (6) $A$ satisfies $r^\prime$.  

\st Statements (4) and (6) imply $A$ satisfies the head of $r^\prime$ and thus the head of $r$ because  $r$ and and $r^\prime$ have the same head.

\st Therefore $r$ is satisfied by $A$, which concludes our
proof of the first part of the proposition.

\st We next prove the second part of the propostion. Consider $p \in A$.  (1) implies that $A$ is an
  answer set of $\alpha(\Pi,A)$. By the supportedness Lemma for ASP
  programs \cite{GelK13}, there is a  rule $r^\prime \in \alpha(\Pi,A)$ such that 

\st (7) $r^\prime$ supports $p$. 

\st Let $r \in \Pi$ be a rule such that $r^\prime = \alpha(r, A)$. By the definition of aggregate reduct, 
  
\st (8) $A$ satisfies the body of $r$ iff 
  $A$ satisfies that of $r^\prime$. 
  
\st Since $r$ and $r^\prime$ have the same heads, (7) and (8) imply that  
  rule $r$ of $\Pi$ supports $p$ in $A$, which concludes the proof of the second part of the proposition.
\hfill $\Box$

\begin{proposition}[Anti-chain Property]
Let $A_1$ be an answer set of an $\mathcal{A}log$ program $\Pi$. 
Then there is no answer set $A_2$ of $\Pi$ such that $A_1$ is a proper 
subset of $A_2$. 
\end{proposition}

\noindent
Proof:  Let us assume that there are $A_1$ and $A_2$ such that

\st
(1) $A_1 \subseteq A_2$ and

\st
(2) $A_1$ and $A_2$ are answer sets of $\Pi$

\st
and show that $A_1=A_2$. 

\st
Let $R_1$ and $R_2$ be the aggregate reducts of $\Pi$ with respect to
$A_1$ and $A_2$ respectively. Let us first show that $A_1$ satisfies the
rules of $R_2$. Consider

\st
(3) $r_2 \in R_2$.

\st
By the definition of aggregate reduct there is $r \in \Pi$ such that 

\st
(4) $r_2 = \alpha(r,A_2)$. 

\st
Consider 

\st
(5) $r_1 = \alpha(r,A_1)$.

\st
If $r$ contains 
no aggregate atoms then 

\st
(6) $r_1 = r_2$.  

\st
By (5) and (6), $r_2 \in R_1$ 
and hence, by (2) $A_1$ satisfies $r_2$.

 \st
Assume now that $r$ contains one aggregate term, $f\{X:p(X)\}$, i.e. $r$ is of the form 

\st 
(7) $h \leftarrow B,C(f\{X:p(X)\})$

\st
where $C$ is some property of the aggregate.

\st 
Then $r_2$ has the form 
 
\st 
(8) $h \leftarrow B,P_2$

\st 
where 

\st 
(9) $P_2 =  \{p(t) : p(t) \in A_2\}$ and $f(P_2)$ satisfies condition $C$.

\st
Let

\st
(10) $P_1 =  \{p(t) : p(t) \in A_1\}$



\st
and consider two cases: 

\st
(11a)  $\alpha(r,A_1) = \emptyset$.

\st
In this case $C(f(P_1))$ does not hold. Hence, $P_1 \not= P_2$.
Since $A_1 \subseteq A_2$ we have that $P_1 \subset P_2$, the body
of rule (8) is not satisfied by $A_1$, and hence the rule (8) is.

\st
(11b) $\alpha(r,A_1) \not= \emptyset$.

\st 
Then $r_1$ has the form 
 
\st 
(12) $h \leftarrow B,P_1$

\st 
where 

\st 
(13) $P_1 =  \{p(t) : p(t) \in A_1\}$ and $f(P_1)$ satisfies condition $C$.

\st
Assume that $A_1$ satisfies the body, $B,P_2$, of rule (8). 
Then 

\st
(14) $P_2 \subseteq A_1$

\st
This, together with (9) and (10) implies

\st
(15) $P_2 \subseteq P_1$.

\st
From (1),
(9), and (10) we have $P_1 \subseteq P_2$.  Hence

\st
(16) $P_1 = P_2$. 

\st
This means that $A_1$ satisfies the body of $r_1$
and hence it satisfies $h$ and, therefore, $r_2$.

\st
Similar argument works for rules containing multiple aggregate atoms
and, therefore, $A_1$ satisfies $R_2$. 

\st
Since $A_2$ is a minimal set satisfying $R_2$ and $A_1$ satisfies
$R_2$ and $A_1 \subseteq A_2$ we have that $A_1 = A_2$.

\st
This completes our proof. \hfill $\Box$

\begin{proposition}[Splitting Set Theorem]
Let 
\begin{enumerate}
\item $\Pi_1$ and $\Pi_2$ be ground programs of $\mathcal{A}log$ such that no atom 
occurring in $\Pi_1$ is unifiable with any atom occurring in the heads 
of $\Pi_2$,
\item  $S$ be a set of ground 
literals containing all head literals of $\Pi_1$ but no head literals of 
$\Pi_2$,
\end{enumerate}
Then 

\st
(3) $A$ is an answer set of $\Pi_1 \cup \Pi_2$

\st
iff

\st
(4a)  $A \cap S$  is an answer set of $\Pi_1$ and

\st
(4b) $A$ is an answer set of $(A \cap S) \cup \Pi_2$. 
\end{proposition}

\st
Proof.  By the definitions of answer set and aggregate reduct 

\st 
(3) holds iff

\st
(5) $A$ is an answer set of $\alpha(\Pi_1,A) \cup \alpha(\Pi_2,A)$

\st
It is easy to see that conditions (1), (2), and the definition of
$\alpha$ imply that $\alpha(\Pi_1,A)$, $\alpha(\Pi_2,A)$, and $S$ satisfy
condition of the splitting set theorem for ASP \cite{LifschitzT94}.
Hence

\st 
(5) holds iff

\st
(6a)  $A \cap S$  is an answer set of $\alpha(\Pi_1,A)$

\st
and

\st
(6b) $A$ is an answer set of $(A \cap S) \cup \alpha(\Pi_2,A)$. 

\st
To complete the proof it suffices to show that  

\st 
(7) Statements (6a) and (6b) hold iff
(4a) and (4b) hold.

\st
By definition of $\alpha$ , 

\st
(8)  $(A \cap S) \cup \alpha(\Pi_2,A) = \alpha((A \cap S) \cup \Pi_2,A)$

\st
and hence, by the definition of answer set we have

\st
(9) (6b) iff (4b).


\st
Now notice that from (4b), clause 2 of Proposition \ref{p1}, and
conditions (1) and (2) of our theorem we have that for any ground instance $p(t)$ of a
literal occurring in an aggregate atom of $\Pi_1$

\st 
(10) $p(t) \in A$ iff $p(t) \in A \cap S$

\st
and, hence

\st
(11) $\alpha(\Pi_1,A) = \alpha(\Pi_1,A \cap S)$.

\st
From (9), (11), and the definition of answer set we have that 

\st
(12) (6a) iff (4a) 

\st
which completes the proof of our theorem. \hfill $\Box$

\begin{lemma} \label{lmm:coNP}
Checking whether a set $M$ of literals is an answer set of $P$, a program with aggregates, is in co-NP. 
\end{lemma}

\noindent Proof: To prove that $M$ is not an answer set of $P$, we first check if $M$ is not a model of the aggregate reduct of $P$, which is in polynomial time. If $M$ is not a model, $M$ is not an answer set of $P$. Otherwise, we guess a set $M'$ of $P$, and check if $M'$ is a model of the aggregate reduct of $P$ and $M' \subset M$. This checking is also in polynomial time. Therefore, the problem of checking whether a set $M$ of literals is an answer set of $P$ is in co-NP. \hfill $\Box$

\begin{proposition}[Complexity]
The problem of checking if a ground atom $a$ belongs to all answer 
sets of an $\mathcal{A}log$ program is $\Pi_2^P$ complete. 
\end{proposition}
\noindent Proof: First we show that the cautious reasoning problem is in $\Pi_2^P$.  
We verify that a ground atom $a$ is not a cautious consequence of a program $P$ as follows: Guess a set $M$ of literals
and check that (1) $M$ is an answer set for $P$, and (2) $a$ is not true wrt $M$. Task (2) is clearly polynomial,
while (1) is in co-NP by virtue of Lemma~\ref{lmm:coNP}. The problem therefore lies in $\Pi_2^P$.

\st Next, cautious reasoning over programs without aggregates is $\Pi_2^P$ hard
by \cite{DantsinEGV01}. Therefore, cautious reasoning over programs with aggregates is $\Pi_2^P$ hard too.

\st In summary, cautious reasoning over programs with aggregates is $\Pi_2^P$ complete. \hfill $\Box$

\end{document}